\documentclass[reqno,11pt]{article}

\usepackage{booktabs}       
\usepackage{amsfonts}       
\usepackage{nicefrac}       
\usepackage[pdftex]{graphicx}
\usepackage{amsmath}
\usepackage{multirow}

\title{\vspace*{-1.5cm} The low-rank hurdle model}

\author{Christopher Dienes\footnote{Seagate Technologies, Operations \& Technology Advanced Analytics Group, 389 Disc Drive, Longmont, CO 80503, USA, email: {\tt crdienes@gmail.com}}} 
\date{\vspace{-5ex}}

\newcommand{\ba}{\mbox{\boldmath $a$}}
\newcommand{\bx}{\mbox{\boldmath $x$}}
\newcommand{\bW}{\mbox{\boldmath $W$}}
\newcommand{\bmu}{\mbox{\boldmath $\mu$}}
\newcommand{\bSigma}{\mbox{\boldmath $\Sigma$}}
\newcommand{\bbeta}{\mbox{\boldmath $\beta$}}
\newcommand{\be}{\mbox{\boldmath $e$}}
\newcommand{\by}{\mbox{\boldmath $y$}}
\newcommand{\bz}{\mbox{\boldmath $z$}}
\newcommand{\bA}{\mbox{\boldmath $A$}}
\newcommand{\bX}{\mbox{\boldmath $X$}}
\newcommand{\bY}{\mbox{\boldmath $Y$}}
\newcommand{\bYj}{\mbox{\boldmath $Y_j$}}
\newcommand{\bYone}{\mbox{\boldmath $Y_1$}}
\newcommand{\bYp}{\mbox{\boldmath $Y_p$}}
\newcommand{\bZ}{\mbox{\boldmath $Z$}}
\newcommand{\bG}{\mbox{\boldmath $G$}}
\newcommand{\bI}{\mbox{\boldmath $I$}}
\DeclareMathOperator*{\argmin}{arg\,min}
\DeclareMathOperator*{\argmax}{arg\,max}

\begin{document}

\maketitle

\begin{abstract}
\noindent A composite loss framework is proposed for low-rank modeling of data consisting of interesting and common values, such as excess zeros or missing values. The methodology is motivated by the generalized low-rank framework and the hurdle method which is commonly used to analyze zero-inflated counts. The model is demonstrated on a manufacturing data set and applied to the problem of missing value imputation.\\

\end{abstract}

\setcounter{equation}{0}
\setcounter{figure}{0}
\setcounter{table}{0}

\section{Introduction}\label{sec:1}

Principal component analysis (PCA) is a popular data science tool used for tasks such as dimensionality reduction, feature extraction, missing value imputation, denoising, and data compression. PCA originated from the works of Pearson \cite{p01} and Hotelling \cite{h33,h36} and a detailed review of the subject is provided by Jolliffe \cite{j86}. Eckart and Young \cite{ey36} described PCA as finding the best approximation of a numeric matrix $\bA$ using a lower rank matrix $\bZ$, where the quality of the approximation is measured using least squares or quadratic loss. 

Numerous authors have extended the concepts of PCA by changing the loss function and adding regularization to the low-rank matrix approximation problem. Notably, Collins et al.\ \cite{c01} proposed using exponential family loss functions and Gordon \cite{g02} used matching link-loss function pairs to construct procedures based on Bregman divergence. These contributions generalized PCA and factor analysis similar to how generalized linear models \cite{mn90} extended the concepts of regression. Regularization has been used to construct low-rank approximations which account for data characteristics such as sparseness \cite{wth09} and non-negativity \cite{ls01}. Udell et al.\ \cite{u16} summarized many of the major contributions using the generalized low-rank model framework. 

This paper is focused on the task of constructing a low-rank approximation when some of the measured variables contain interesting values which occur frequently. Examples include missing, censored, or truncated values; as well as zero-inflated data. The zero-inflated case is commonly encountered when measuring manufacturing defect counts. For regression analysis settings, Mullahy \cite{m86} proposed using the hurdle model, Lambert \cite{l92} described the zero-inflated model, and Min and Agresti \cite{ma05} provided enhancements which included random effects. For dimensionality reduction, Pierson and Yau \cite{py15} developed the zero-inflated factor analysis (ZIFA) model for analyzing single cell RNA sequencing data suffering from gene expression dropout. The ZIFA model follows the probabilistic PCA approach of Tipping and Bishop \cite{tb99} and optimization is carried out via the EM algorithm \cite{dlr77}. The ZIFA model can be expressed as a special case of the low-rank reduced hurdle model presented in Section \ref{sec:3}. The case of performing PCA in the presence of missing data has been examined previously, with Ilin and Raiko \cite{ir10} providing a review of existing procedures. The low-rank hurdle model offers a new representation which can be leveraged to gain additional data insights not directly available from competing PCA missing data methods.       

The remaining contents are organized as follows. Section \ref{sec:2} describes the generalized low-rank framework. The hurdle model is motivation in Section \ref{sec:3}, along with details for proper implementation. In Section \ref{sec:4} the hurdle approach is used to analyze a zero-inflated manufacturing data set and investigate missing value imputation. Lastly, Section \ref{sec:5} contains some concluding remarks.


\section{The generalized low-rank model}\label{sec:2}

The following notation is used throughout. Matrices are denoted by bold uppercase letters or Greek symbols (e.g.\ $\bA$, $\bSigma$) , vectors are represented by bold lowercase letters or Greek symbols (e.g.\ $\ba$, $\bmu$), and scalars are not bold (e.g.\ $a_{ij}$, $\mu_j$). Additionally, matrices with dimensions $m \times d_j$ and vectors of length $d_j$ are denoted as matrices and vectors; respectively, even if $d_j = 1$ occurs for some $j$.   

Here we present a generalized framework for low-rank modeling, summarizing the methodology highlighted by Udell et al.\ \cite{u16}. Let $\bA$ be an $n \times p$ data table where the rows represent $n$ observations consisting of measurements collected on $p$ variables. Then for $i = 1,\ldots,n$ and $j=1,\ldots,p$, $a_{ij}$ represents the $j^{th}$ variable value for the $i^{th}$ observation. The domain for each column variable is denoted by $\mathcal{F}_j$, which is not restricted to $\mathbb{R}$, but includes discrete and non-numeric domains to facilitate abstract data types such as count, Boolean, categorical, and ordinal variables. We will approximate abstract data types by representing $a_{ij} \in \mathcal{F}_j$ with numerical embeddings $\bz_{ij} \in \mathbb{R}^{d_j}$, where $d_j$ is the embedding dimension of the $j^{th}$ variable. The resulting embedded dimension of the model is $d = \sum_j d_j$. The loss incurred from using $\bz_{ij}$ to describe $a_{ij}$ is measured using an appropriately selected loss function $L_{ij} : \mathbb{R}^{d_j} \times \mathcal{F}_j \rightarrow [0, \infty)$.

Essential to this analysis is the construction of a low-rank matrix $\bZ \in \mathbb{R}^{n \times d}$ which approximates our data table with minimal loss. A rank-$k$ approximation can be found by specifying $\bZ = \bX\bY$ where $k < d$, $\bX \in \mathbb{R}^{n \times k}$, and $\bY \in \mathbb{R}^{k \times d}$. Notice this decomposition is not unique since $\bZ = \bX\bY = \bX\bG^{-1}\bG\bY$ for any non-singular $k \times k$ matrix $\bG$. An optimal rank-$k$ matrix decomposition can be found by minimizing the following optimization problem. Let $\bx_i \in \mathbb{R}^{1 \times k}$ denote the $i^{th}$ row of $\bX$, and $\bY = \left[\bYone \cdots \bYp\right]$ such that $\bYj \in \mathbb{R}^{k \times d_j}$ denotes the embedded columns associated with the $j^{th}$ variable of $\bA$, then the \emph{generalized low-rank model} for $\bA$ is found using 
\begin{equation}\label{eq:2.1}
   \mbox{minimize } \sum_{(i,j)\in \Omega} L_{ij}(\bx_i \bYj, a_{ij}) + \sum_i r_i(\bx_i) + \sum_j \tilde{r}_j(\bYj),
\end{equation}
where $r_i : \mathbb{R}^{1\times k} \rightarrow [0, \infty)$ and $\tilde{r}_j : \mathbb{R}^{k\times d_j} \rightarrow [0, \infty)$ are appropriately selected regularizers, and $\Omega \subseteq \{1,\ldots,n\}\times\{1,\ldots,p\}$ represents the set of indices $(i,j)$ such that $a_{ij}$ is observed. An appealing feature of the above generalized structure is the ability to combine different loss functions and regularizers to address different variable characteristics observed in the data table. 

Many data reduction methods can be described in terms of equation (\ref{eq:2.1}). For example, if unregularized quadratic loss is chosen for a numeric data table $\bA$ with no missing values, then the optimization problem is solved using standard PCA \cite{ey36}. This motivates the interpretation of the matrix $\bX$ as a low-dimensional representation of $\bA$, with $\bY$ representing a mapping of $\bX$ back into the original embedded data space. Other special cases described by the general framework include robust and sparse PCA, exponential family PCA, non-negative matrix factorization, and matrix completion \cite{u16}.

The task of optimizing equation (\ref{eq:2.1}) is simplified for convex loss functions and regularizers. Under these conditions (\ref{eq:2.1}) becomes a biconvex minimization problem, which is commonly solved iteratively by alternating between convex updates in one argument while fixing the other. Using the above notation, we alternate minimization over the rows of $\bX$ while fixing $\bY$, and minimization over the columns of $\bY$ while fixing $\bX$. These updates can be parallelized over the rows of $\bX$ and the columns of $\bY$ which may significantly improve computing times. In general this alternating approach does not guarantee convergence to the global minimizer, and care may be required to avoid poor solutions. In many applications, the usefulness of the sub-optimal solution is used to justify its adoption.   

Variable scaling is a well known issue in multivariate analysis, and commonly data is normalized prior to performing methods such as PCA. The concepts of offset and scaling can be generalized by replacing the loss functions in (\ref{eq:2.1}) by $L_{ij}(\bx_i \bYj + \bmu_j, a_{ij})/\sigma^2_j$, where
\begin{equation}\label{eq:2.2}
    \bmu_j = \argmin_{\bmu \in \mathbb{R}^{d_j}} \sum_{(i,j)\in \Omega} L_{ij}(\bmu, a_{ij}), \quad \quad \sigma^2_j = \frac{1}{n_j - 1} \sum_{(i,j)\in \Omega} L_{ij}(\bmu_j, a_{ij}),
\end{equation}
and $n_j$ is the number of non-missing values for the $j^{th}$ variable. Using the above expressions, the loss contribution for the $j^{th}$ variable is equal to $n_j - 1$ under the offset only model. This motivates the use of $\sum_j (n_j - 1)$ as the total loss of the scaled model, which has a similar interpretation as total variation from standard PCA analysis. It is important to note the offset and scaling adjustments are applied to the loss functions, and not directly to the data table itself. This ensures aspects of the data table are maintained, such as sparseness or non-negativity.


\section{The hurdle model}\label{sec:3}

Suppose data table $\bA$ contains variable $\ba_j$ with elements $a_{ij} \in \mathcal{F}_j$ which periodically take on the value $\nu \in \mathcal{F}_j$. Assume the occurrence of $a_{ij} = \nu$ is interesting because it potentially signifies a different generating process as compared to when $a_{ij} \neq \nu$. For example, defect counts are observed during the manufacturing of hard disc drives. Normal counts are typically zero and appear to be governed by a process which differs from non-zero defect counts; where differences are observable across the data table variables. Regression analysis techniques have been proposed for this paradigm, including the hurdle model from Mullahy \cite{m86} and the zero-inflated model from Lambert \cite{l92}. 

The hurdle model contains two components, where the first component represents the probability of observing $\nu$ and the second describes the conditional behavior of the data provided $\nu$ is not observed. Explicitly,  
\begin{align*}
    \Pr[a_{ij} = \nu] & =  p_{ij},\\
    f_{a_j/\nu}(a_{ij}\,; \mu_{ij}) & = (1 - p_{ij})g(a_{ij}\,; \mu_{ij}) \quad \mbox{for } a_{ij} \neq \nu,
\end{align*}
where $f_{a_j/\nu}(\cdot \, ; \mu_{ij})$ represents the probability density or mass function when $\nu$ is not observed, and $g(\cdot \, ; \mu_{ij})$ is the possibly $\nu$-truncated density or mass function with mean parameter $\mu_{ij}$. Following the generalized linear model framework \cite{mn90}, appropriate mean functions $(\eta_1,\eta_2)$ can be defined such that 
\[p_{ij} = \eta_1(\bx_{i1}\bbeta_1) \quad \mbox{and} \quad \mu_{ij} = \eta_2(\bx_{i2}\bbeta_2),\]
where $(\bx_{i1}, \bx_{i2})$ represent predictor row vectors and $(\bbeta_1, \bbeta_2)$ represent parameter column vectors. Under the typical assumption of logit link for the probabilities $p_{ij}$, maximum likelihood estimation is performed using the following equation:
\begin{equation}\label{eq:3.1}
    \argmax_{\bbeta_1, \bbeta_2} \, \prod_{i=1}^{n}\left[\frac{\exp(\mathbb{I}{(a_{ij} = \nu)}\bx_{i1}\bbeta_1)}{1+\exp(\bx_{i1}\bbeta_1)}\,\,g\left(a_{ij}\,; \eta_2(\bx_{i2}\bbeta_2)\right)^{1 - \mathbb{I}{(a_{ij} = \nu)}}\right].
\end{equation}
Equation (\ref{eq:3.1}) can be expressed as a minimization problem by examining the negative log of the likelihood function, which yields
\begin{equation}\label{eq:3.2}
    \argmin_{\bbeta_1, \bbeta_2} \, \sum_{i=1}^{n}\log\left[1 + \exp(-a_{ij}^*\bx_{i1}\bbeta_1)\right] - \sum_{i:a_{ij}\neq\nu} \log\left[g\left(a_{ij}\,; \eta_2(\bx_{i2}\bbeta_2)\right)\right],
\end{equation}
where $a_{ij}^* = 2*\mathbb{I}{(a_{ij} = \nu) - 1}$ is an embedded indicator variable. Previous authors have examined low-rank procedures motivated by loss functions based on the negative log likelihood; notably Collins et al.\ \cite{c01} in the case of exponential family models. In the context of the generalized low-rank model presented earlier (\ref{eq:2.1}), denoting $\bYj = (\by_{j,1}, \by_{j,2}) \in \mathbb{R}^{k \times 2}$ and replacing $(\bx_{i1}\bbeta_1,\bx_{i2}\bbeta_2)$ with $(\bx_{i}\by_{j,1} + \mu_{j,1},\bx_{i}\by_{j,2} + \mu_{j,2})$ in (\ref{eq:3.2}) yields the following equivalent low-rank expression for fixed $j$:
\begin{align}
  \sum_{i=1}^{n}L_{ij}(\bx_i \bYj + \bmu_j, a_{ij}) &= \sum_{i=1}^{n}\log\left[1+\exp(-a_{ij}^{*}(\bx_{i}\by_{j,1}+\mu_{j,1}))\right]\label{eq:3.3}\\ 
  & \quad \quad \quad + \sum_{i:a_{ij}\neq\nu} \log\left[\frac{g\left(a_{ij}\,; \eta_2(m_{ij})\right)}{g\left(a_{ij}\,; \eta_2(\bx_{i}\by_{j,2} + \mu_{j,2})\right)}\right] \\ 
  = \sum_{i=1}^{n}L_{\ell,ij}(\bx_{i}\by_{j,1} & + \mu_{j,1}, a_{ij}^{*}) + \mathbb{I}{(a_{ij} \neq \nu)}L_{g,ij}(\bx_{i}\by_{j,2} + \mu_{j,2},a_{ij}),\label{eq:3.5}
\end{align}
where $L_{\ell,ij}$ denotes logistic loss, $L_{g,ij}$ represents a $g(\cdot;\cdot)$ derived loss, and $m_{ij} = \argmax_c g\left(a_{ij}\,; \eta_2(c)\right)$ is a normalizing constant to ensure $L_{g,ij}$ is non-negative. The derived equation in (\ref{eq:3.5}) can be further generalized for arbitrary data structures by making use of the subsequent composite loss definitions.

{\bf Definition} (hurdle loss). Let $\bz = (z_1, z_2) \in \mathbb{R}^{2}$, $a \in \mathcal{F}$, $\nu \in \mathcal{F}$, $\lambda_1, \lambda_2 > 0$, and $a^{*}$ be a binary variable indicating whether $a = \nu$ has occurred, then \emph{full hurdle loss} $L_{fh} : \mathbb{R}^{2} \times \mathcal{F} \rightarrow [0, \infty)$ is specified by
\begin{equation}\label{eq:fhl}
   L_{fh}(\bz, a)  = \lambda_1 L_b(z_1, a^{*}) + \mathbb{I}{(a \neq \nu)}\lambda_2 L_g(z_2, a),
\end{equation}
where $L_b$ denotes a non-negative binary loss, and $L_g$ is an appropriate non-negative loss for describing the $\nu$-truncated data. Furthermore let $z \in \mathbb{R}$, then \emph{reduced hurdle loss} $L_{rh} : \mathbb{R} \times \mathcal{F} \rightarrow [0, \infty)$ is defined as
\begin{equation}\label{eq:rhl}
   L_{rh}(z, a)  = \lambda_1 L_b(z, a^{*}) + \mathbb{I}{(a \neq \nu)}\lambda_2 L_g(z, a).
\end{equation}

The weights $\lambda_{1}, \lambda_{2}$ assign relative importance to the two model components, with larger weights implying higher importance on the resulting reduced representation. The choice of weights will also affect the aggregated total loss, but this can be corrected using the formulas for offset and scaling appearing in equations (\ref{eq:2.2}). One strategy is to assign weights proportional to total loss contributions resulting from the two hurdle components, where total loss is found using the offset only model. Specifically  for the $j^{th}$ variable, allow $n_{j,\nu}$ to represent the number of $\nu$ occurrences, $n_j - n_{j,\nu}$ the number of non-$\nu$ occurrences, and offsets $\mu_b$ and $\mu_g$ are found using (\ref{eq:2.2}), then weights $\lambda_{j,1}, \lambda_{j,2}$ which solve the below system of equations will yield a total loss of $n_j -1$ and ensure the binary loss contributes $c$ times the non-$\nu$ loss:
\begin{equation}
\begin{bmatrix}
1 & 1\\
1 & -c\\
\end{bmatrix}
\begin{bmatrix}
\sum_{i \in \Omega} L_{b,ij}(\mu_b, a_{ij}^*) & 0\\
0 & \sum_{i \in \Omega} \mathbb{I}{(a_{ij} \neq \nu)} L_{g,ij}(\mu_g, a_{ij})
\end{bmatrix}
\begin{bmatrix}
\lambda_{j,1}\\
\lambda_{j,2}
\end{bmatrix}
=
\begin{bmatrix}
n_j - 1\\
0
\end{bmatrix}\label{eq:lambdas}
\end{equation}
where $c$, $\lambda_{j,1}$, $\lambda_{j,2} > 0$. Several intuitive choices for the multiplier are $c = 1$ or $c = n_{j,\nu} / (n_j - n_{j,\nu})$.

In practice, data analysts are required to make several decisions in order for hurdle loss to be implemented within the generalized low-rank framework described in Section \ref{sec:2}. Logistic loss is likely a default choice for the binary loss $L_b$, while selecting a form for $L_g$ may depend more heavily on the underlying data characteristics. For example, quadratic and $\ell_1$ losses are reasonable choices for continuous data, while Poisson loss is useful for count variables. As previously mentioned, selecting convex loss functions and regularizers allows alternating minimization to be employed which may further guide the decision. A more detailed overview of possible loss functions and regularizers is provided in Udell et al.\ \cite{u16}. 

Reduced hurdle loss (\ref{eq:rhl}) simplifies the representation by setting $\by_{j,1} = \by_{j,2} = \by_j$ and $\mu_{j,1} = \mu_{j,2} = \mu_j$ in (\ref{eq:3.5}). In the special case of the ZIFA model \cite{py15}, quadratic loss is selected for $L_g$ and $L_b$ is based on the binomial probabilities $p_{ij} = \exp(-\xi_j (\bx_i \by_j + \mu_j)^2)$ where $\xi_j$ is a positive decay coefficient. The ZIFA model substitutions were justified in the context of the gene expression data problem and may be unsuitable for other subject domains.    

In applications such as matrix completion and data reconstruction, mapping the reduced representation back into the original domain of $a_{ij}\in\mathcal{F}_j$ is required. In general, low-rank models approximate $a_{ij}$ using some function of the vector $\bz_{ij} = \bx_i\bYj + \bmu_j$. For quadratic loss this is simply $\hat{a}_{ij} = z_{ij}$ since this model corresponds to the generalized linear model with identity link.  Under hurdle loss the original variable is encoded using both an indicator function and the identity function, where the latter function is applied only when non-$\nu$ values occur. These encodings are then approximated using a vector $\bz \in \mathbb{R}^{2}$ in the full model setting, or $z \in \mathbb{R}$ for reduced models such as ZIFA. Hence, the reverse mapping under hurdle loss can be found as follows. Let $\tilde{a}_{ij} = \argmin_{a\in\mathcal{F}_j/\nu}L_{g,ij}(\bx_i\by_{j,2} + \mu_{j,2}, a)$ and $a_{ij}^* = \mathbb{I}{(a_{ij} = \nu)}$, then the reconstructed value $\hat{a}_{ij}$ is determined by        
\begin{align}
  \hat{a}_{ij} &= \argmin_a L_{h,ij}(\bx_i\bYj + \bmu_j, a)\\
               &= \begin{cases} \nu  & \mbox{if}\quad \dfrac{\lambda_{j,1} L_{b,ij}(\bx_i\by_{j,1} + \mu_{j,1}, 0) + \lambda_{j,2} L_{g,ij}(\bx_i\by_{j,2} + \mu_{j,2}, \tilde{a}_{ij})}{\lambda_{j,1} L_{b,ij}(\bx_i\by_{j,1} + \mu_{j,1}, 1)} >1 ,\label{cond:2}\\
               \tilde{a}_{ij} & \mbox{otherwise.}
\end{cases}
\end{align}
The above piecewise condition is often simplified since $\tilde{a}_{ij} = \bx_i\by_{j,2} + \mu_{j,2}$ and $L_{g}(z,z) = 0$ for commonly used loss functions, such as quadratic and Poisson losses. Hence the condition in (\ref{cond:2}) becomes a ratio of only the binary loss components. In the case of missing data, $\tilde{a}_{ij}$ is still a reasonable imputed value even when the binary loss components suggest missingness is likely. This is further demonstrated in Subsection \ref{sec:4b}.   

The foundations for hurdle loss were developed following a likelihood model explanation, but additional intuition and advantages are worth mentioning. First, in the full model setting each of the hurdle components receives a different principal vector in $\bYj = (\by_{j,1},\by_{j,2})$ which allows for differing dependencies on the other columns of $\bA$. This added flexibility mirrors the modeling complexities available in the hurdle and zero-inflated regression frameworks. Low-rank applications concerned with data similar to those which motivated the regression models may find using the hurdle approach a suitable alternative to competing dimension reduction methods. Second, since the low-dimensional representation retains information related to the likelihood of $\nu$ values, we may extract probability type scores using $1/\left[1 + \exp(- \bx_i\by_{j,1} - \mu_{j,1})\right]$ and measure associations with other variables by examining the cosine similarity between $\by_{j,1}$ and the remaining columns of $\bY$. These metrics inform the analyst about the quality of the low-rank representation with respect to discriminating $\nu$ values, without the need to conduct additional analysis. Lastly, employing the composite hurdle loss provides an additional degree of freedom when determining the offset and scaling for the underlying variable. These values can be strongly influenced by $\nu$-inflation when using a single loss function, potentially obscuring meaningful representations of the underlying processes.


\section{Applications}\label{sec:4}

\subsection{Zero-inflated model}\label{sec:4a}

The first example investigates a factory data set which contains various defect count variables related to the manufacturing of hard disc drives. In general, defect count variables tend to exhibit high degrees of zero-inflation. This particular data set contains 14 different count variables measured on 2200 unique storage devices. The observed zero-inflation varies across variables and ranges from $5\%$ to $99\%$, with an aggregated value of about $60\%$. The distribution of non-zero values displays a long tail with an overall median and mean of $2$ and $13.3$ defects, respectively. 

The generalized low-rank model (\ref{eq:2.1}) was used to analyze the defect data set. Unregularized hurdle loss was chosen for all 14 count variables, with binary and non-zero components selected to be the following logistic and Poisson loss functions
\begin{align*}
   L_\ell(z, a^{*})  &= \log\left[1+\exp(-a^{*}z)\right],\\
   L_p(z, a)     &= \exp(z) -az + a \log(a) - a,
\end{align*}
where $a^* = 2*\mathbb{I}{(a = 0) - 1}$ as before. Note that the likelihood motivated expressions from (\ref{eq:3.3} - \ref{eq:3.5}) would suggest using the loss function derived from the zero-truncated Poisson
\begin{equation*}
L_{tp}(z,a) = \log\left[\exp(\exp(z)) - 1\right] - az + a\log[g(a)] - \log\left[\exp(g(a)) - 1\right],
\end{equation*}
where $g(a) = \argmax_c a\log(c) - \log[\exp(c) - 1]$. However, convergence tends to be slower and numerically unstable when employing the truncated version. Additionally, differences between the ordinary Poisson loss and the zero-truncated version converge quickly to zero as $(z,a)$ increase.  

All loss functions were centered and scaled according to (\ref{eq:2.2}). Specifically, offset terms for logistic and Poisson losses are $\log\left[n_{j,\nu} / (n_j - n_{j,\nu})\right]$ and $\log\left(\bar{\ba}_j\right)$, where $\bar{\ba}_j$ is the sample column mean. Additionally, hurdle loss components were weighted and scaled using (\ref{eq:lambdas}) with $c = n_{j,\nu} / (n_j - n_{j,\nu})$. Ordinary PCA and the ZIFA model were also considered for comparative purposes. For the ZIFA model, initial values were based on PCA and the decay parameter $\lambda_j$ was allowed to vary across variables. 
\begin{figure}[t]
  \centering
  \includegraphics[width=0.32\linewidth]{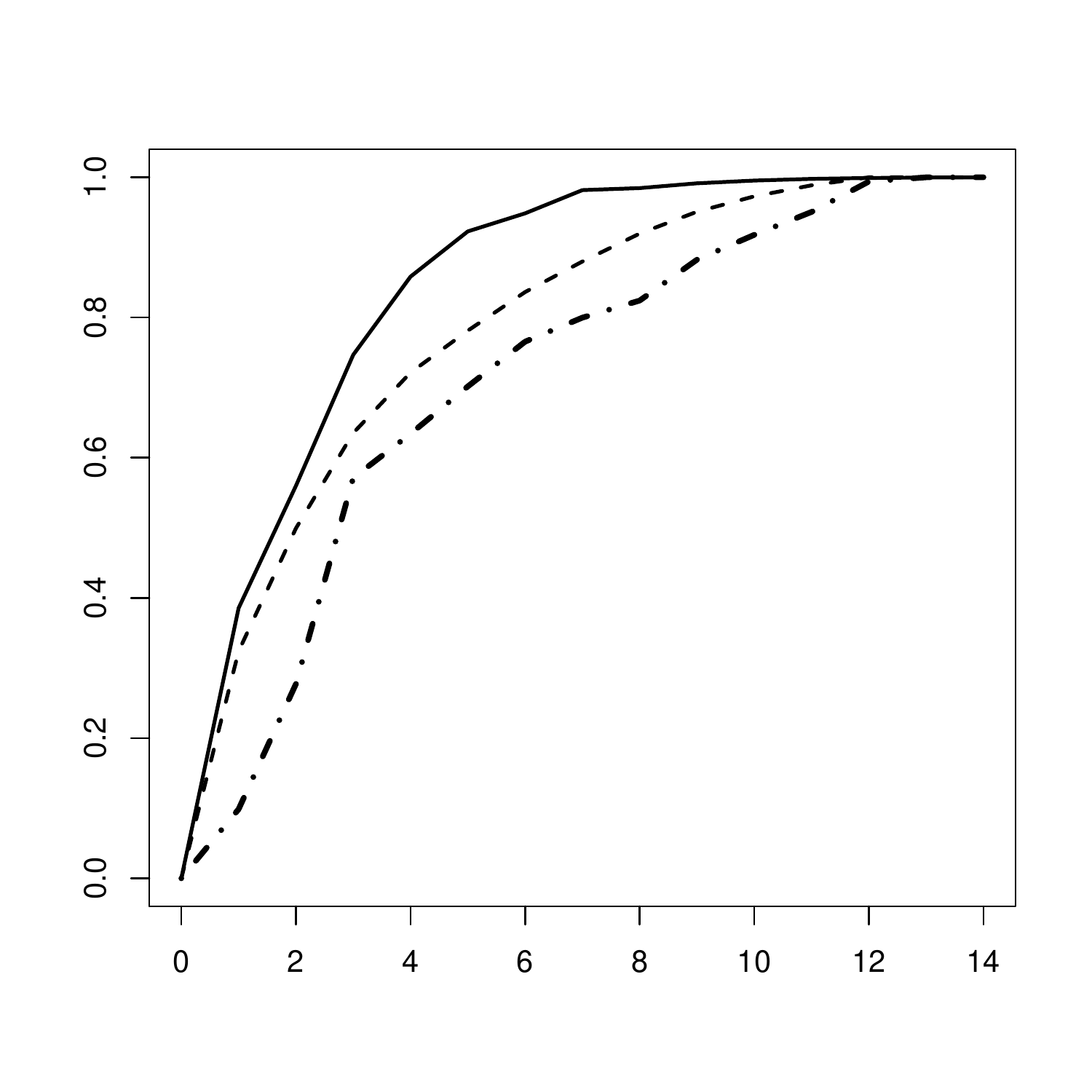}
  \includegraphics[width=0.32\linewidth]{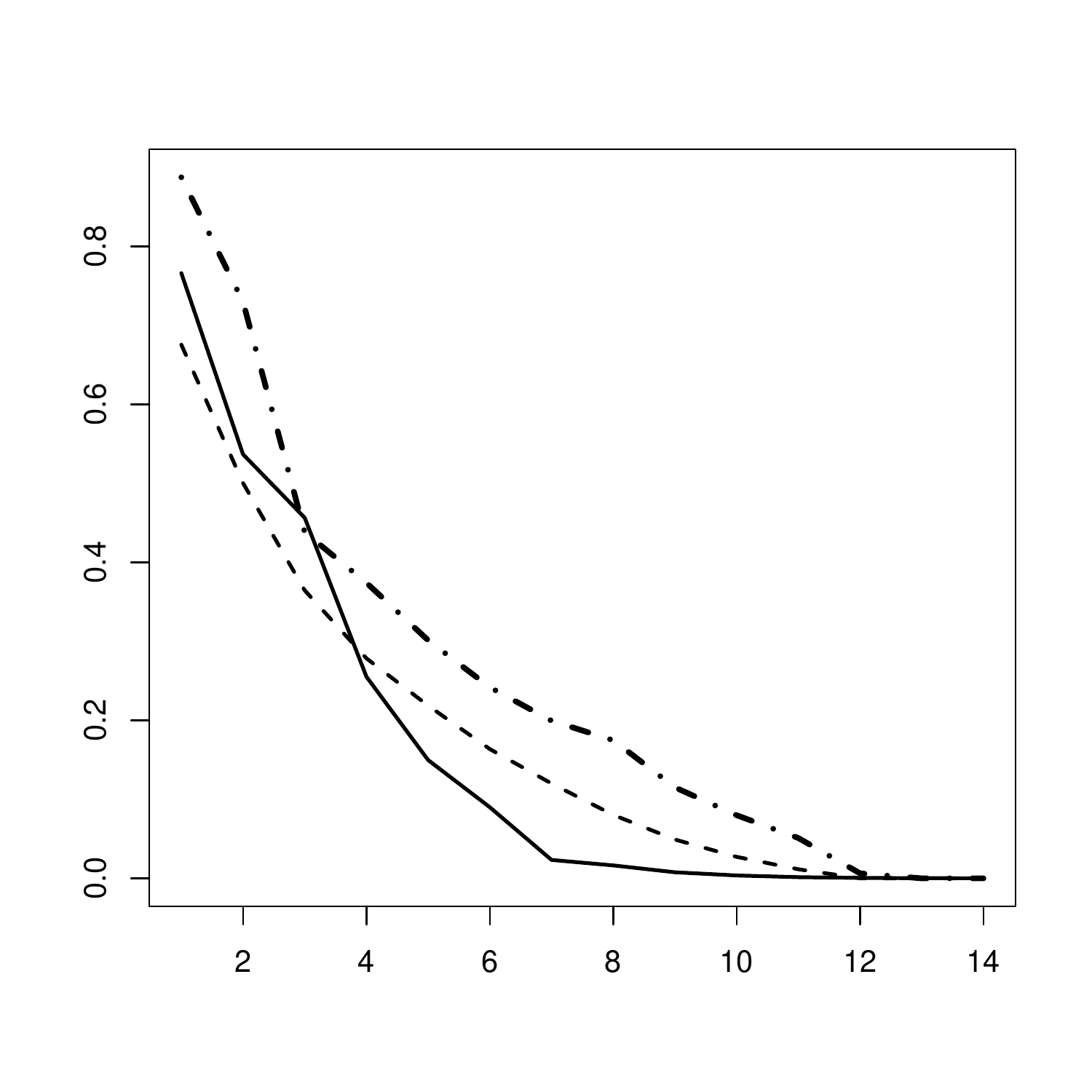}
  \includegraphics[width=0.32\linewidth]{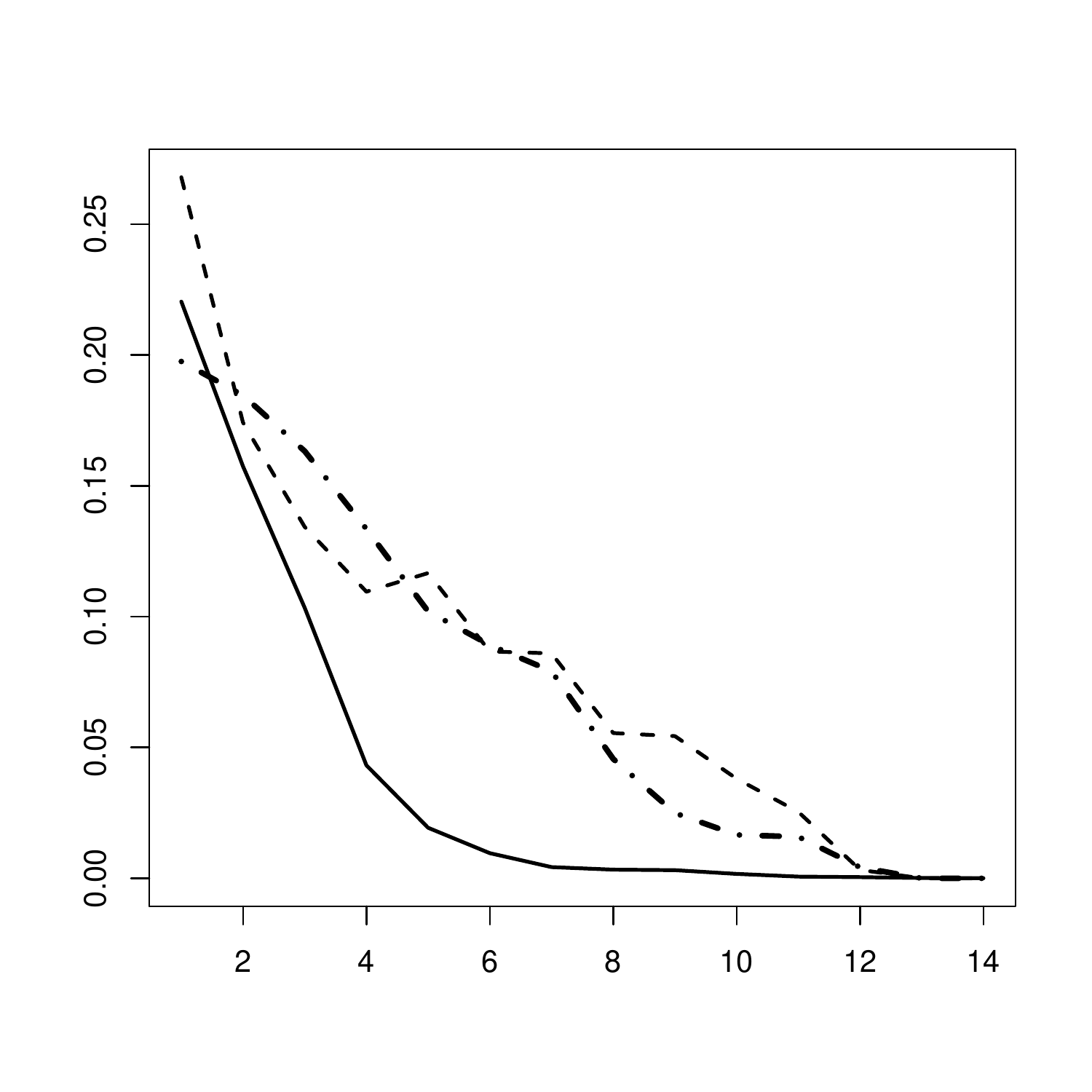}
  \caption{Solid lines depict the hurdle model, dashed lines represent PCA, and dotted-dashed lines denote ZIFA. Left: proportion of loss explained. Middle: weighted reconstruction SSE. Right: zero misclassification rate.}
  \label{fig:zero}
\end{figure}

Figure \ref{fig:zero} contains three plots comparing the full hurdle, ZIFA, and PCA approaches. The left plot displays proportion of total loss explained as a function of the model dimension $k$. Recall total loss is calculated under the offset only model and equals $\sum_j (n_j - 1)$ when the loss functions are appropriately scaled. The hurdle model achieves a quicker rate of model loss reduction with respect to its model loss space, followed by PCA. The middle plot compares element-wise weighted sum of squared reconstruction errors, where the weights are the sample standard deviations of the target variables. The hurdle model performs similarly to PCA and shows improvement over dimensions 4 through 11. The right plot displays zero misclassification rates for the three methods. Simple threshold decision rules are used to map the reduced rank representations into zero/non-zero responses. Specifically, PCA reports a zero outcome whenever a reconstructed value is less than 0.5, while the hurdle and ZIFA models assign a zero value whenever a reconstructed probability score exceeds 0.5. The plot shows the full hurdle model performs noticeable better than PCA and ZIFA, which both performed similarly. Overall, the ZIFA model either performs similar or worse than PCA. The full hurdle model framework includes an additional column in the representation $\bYj$, which may provide advantages when optimizing the potential trade-offs between the competing composite losses. This added flexibility is absent in reduced hurdle models and may explain the degraded ZIFA performance on this data set.   

Missing in the analysis of Figure \ref{fig:zero} is the computational speed advantages of ordinary PCA. A parallelized alternating second order gradient descent procedure was used to fit the hurdle model, and the EM algorithm was used for ZIFA. In general, optimizing the generalized model is slower than ordinary PCA and care needs to be taken to avoid poor local minimums. For applications which require fast implementations, stable representations for $\bY$ can be found offline and held fixed for efficient scoring of new data.


\subsection{Missing value model}\label{sec:4b}

Performing PCA in the presence of missing values is a well studied problem. Ilin and Raiko \cite{ir10} provide a review of common practical approaches to PCA with incomplete data. For our purposes the problem can be reformulated in the context of the hurdle model. Specifically, assume a logistic loss for the occurrence of missingness and quadratic loss for the observed data. This approach is investigated by simulating 30 data sets each containing 5000 observations and 10 variables. Each $10 \times 1$ observation vector $\ba_i$ is generated using the following low-rank sampling scheme:
\begin{align*}
\bz_i &\sim \mbox{N}_4 (0, \bI_4),\\
\be_i &\sim \mbox{N}_{10} (0, \bSigma)\\
\bmu & = \left(1,2,3,4,5,6,7,8,9,10\right)^T\\
\ba_i &= \bW \bz_i + \bmu + \be_i,
\end{align*}
where for each data set $\bSigma$ is a diagonal matrix sampled uniformly from (0.9, 1.1), and $\bW$ is a $10 \times 4$ matrix with entries $w_{k\ell}$ generated from a standard normal distribution. Missingness is induced using two alternative methods applied to the same generated data set. The first approach assumes data is missing completely at random (MCAR), where as the second assumes data is missing at random (MAR) by correlating selection with the observed data. Only the first entries $a_{i1}$ in $\ba_i$ suffer from missingness with exclusions based on the following selection probabilities:
\begin{align*}
\mbox{(MCAR)}\quad\quad\mbox{Pr}\left[a_{i1} \mbox{ is missing} \right] & = \left[1 +\exp(1.7)\right]^{-1},\\
\mbox{(MAR)}\quad\quad\mbox{Pr}\left[a_{i1} \mbox{ is missing} \right] & = \left[1 +\exp(\alpha + a_{i2} + a_{i3})\right]^{-1}.
\end{align*}
The value of $\alpha$ is recalculated for each data set so that the rate of missingness is approximately the same under both MCAR and MAR cases; yet under MAR, missingness is directly associated with the observed values of the second and third measured variables.  

Under the zero-inflated model, offset terms for the $\nu$-truncated loss were found using (\ref{eq:2.2}). For quadratic loss this suggests using the sample mean. However under missing data the sample mean is known to be a biased estimate for the offset term \cite{ir10}, especially when considering MAR type missingness. To account for bias, the offset term for the first variable was updated between alternating minimization steps using
\[\mu_{1,2} = \frac{1}{n_1} \sum_{i\in\Omega}(a_{i1} - \bx_i\by_{1,2}).\]
Scaling for the first variable's hurdle loss components followed (\ref{eq:lambdas}) with $c = n_{1,\nu} / (n_1 - n_{1,\nu})$. The remaining nine variables were modeled using only quadratic loss with offset and scaling terms found using (\ref{eq:2.2}).

Regularization was included in the low-rank model to reduce over-fitting and improve data imputation. Quadratic regularizers $r(\bx) = \gamma_x ||\bx||^2_2$ and $\tilde{r}(\by) = \gamma_y ||\by||^2_2$ were selected, and for simplicity $\gamma = \gamma_x = \gamma_y$ was assumed. In order to choose the regularization parameter $\gamma$, missing data was omitted from the generated data sets and new missing values were randomly created using a MCAR scheme with a similar selection rate as observed in the generated data. The new missing values were imputed using a range of $\gamma$ values and the mean squared imputation error was used to find optimal values.    

\begin{figure}[t]
  \centering
  \includegraphics[width=0.32\linewidth]{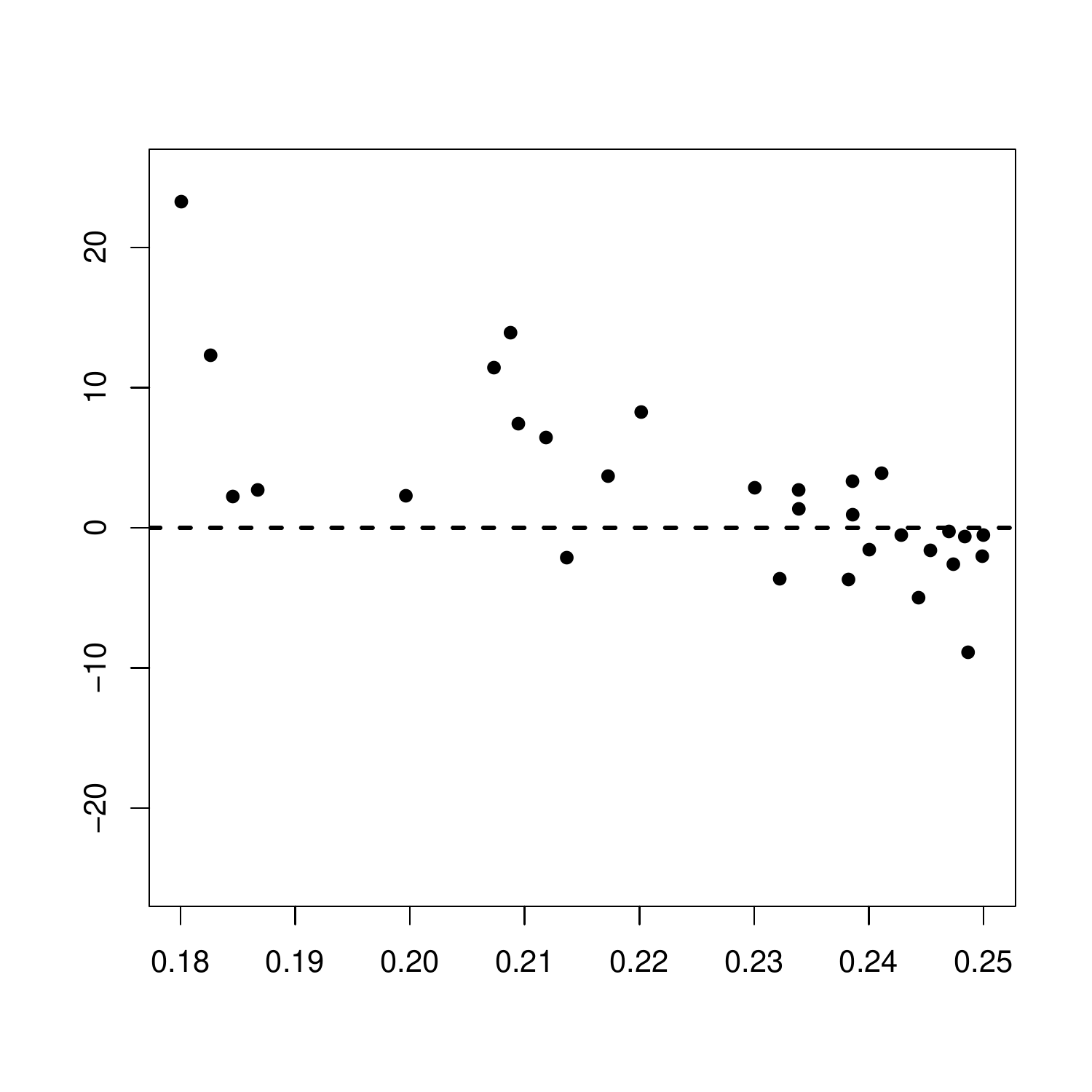}
  \includegraphics[width=0.32\linewidth]{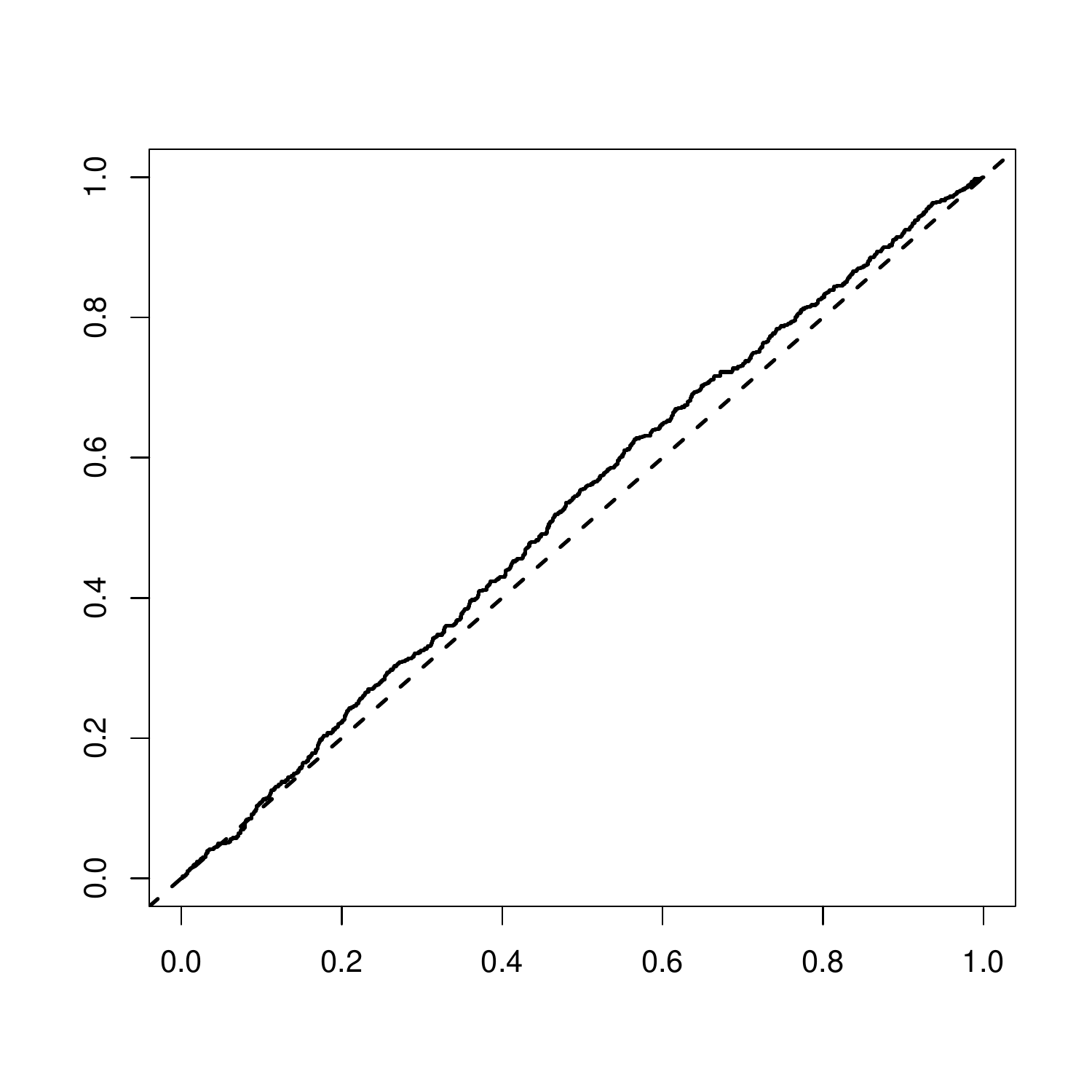}
  \includegraphics[width=0.32\linewidth]{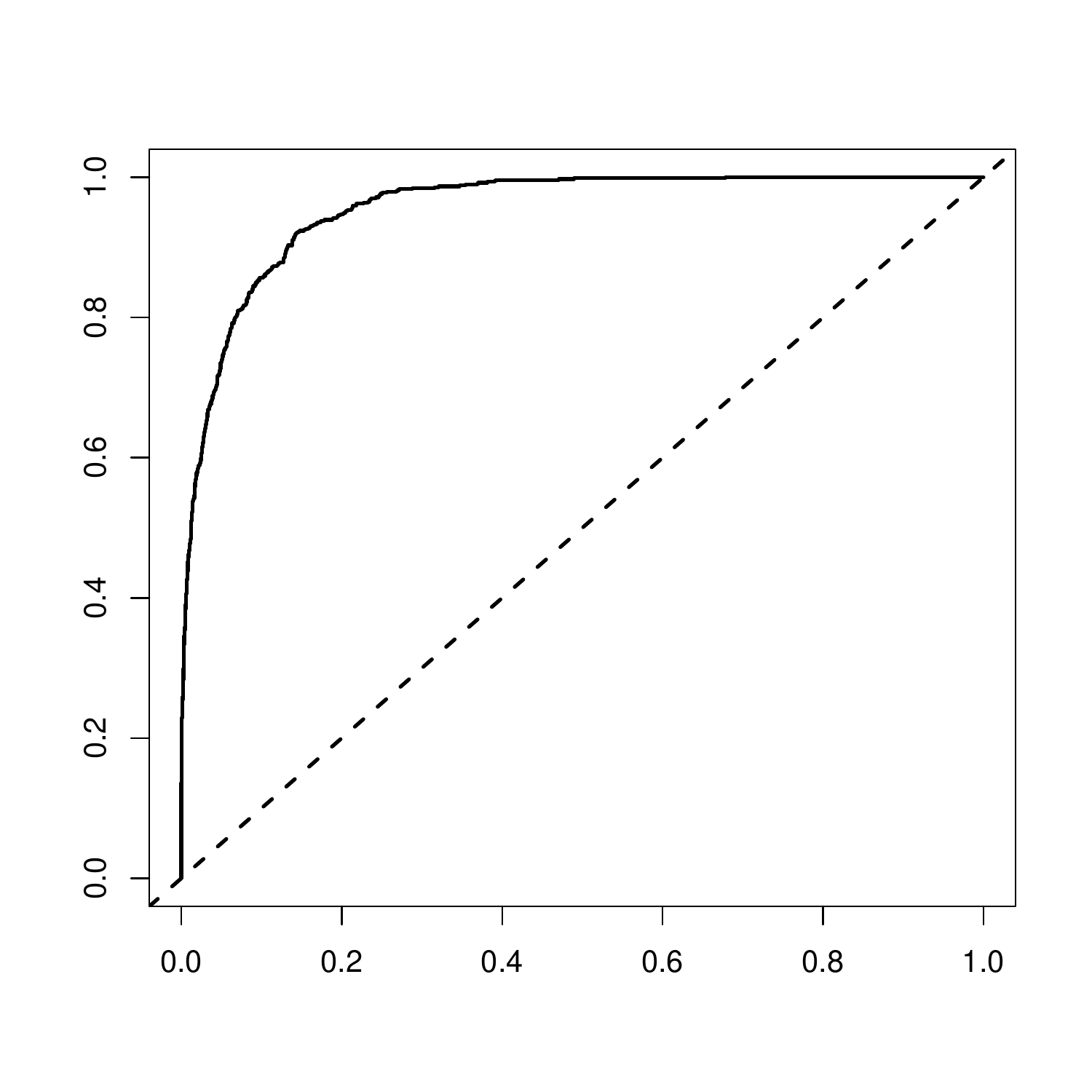}
  \caption{Left: hurdle model MSE improvement plotted against data separation. Middle: hurdle ROC curve for the first MCAR data set. Right: hurdle ROC curve for the first MAR data set.}
\label{fig:na}
\end{figure}

The regularized full hurdle model was applied to the MCAR and MAR data sets, along with four additional models for comparative purposes: Bayesian PCA (BPCA) \cite{o03}, Probabilistic PCA (PPCA) \cite{r98}, Nonlinear Iterative Partial Least Squares (NIPALS) \cite{w66}, and imputation using the sample mean of the observed values. All the data reduction techniques assumed a $k=4$ reduced representation. The performance of the various methods was measured based on imputation and offset mean squared errors, and average performance is reported in Table \ref{tab:impute}. In both cases the low-rank models significantly improve upon the sample mean approach. In the MCAR setting, adding the hurdle structure is unnecessary causing performance to be slightly worse than the BPCA and PPCA approaches. The hurdle model reports the overall best performance for the MAR data sets. Under the MAR setting, missing values provide additional information regarding the underlying data structure which the hurdle model more accurately represents. This point is further expressed by the left most plot in Figure \ref{fig:na}. For each of the MAR data sets, the probability $\rho$ of the observed $a_{i1}$ values exceeding the unobserved $a_{i1}$ missing values was recorded. Data sets with high separation between observed and missing distributions exhibit small $\rho (1-\rho)$ values. This separation measure was compared to the percentage improvement in MSE for the hurdle model over BPCA, where positive values indicate better performance for the hurdle model. The plot clearly reveals the hurdle model becomes more preferable as the overlap between the observed and missing data decreases.             

\begin{table}[t]
\caption{Missing data imputation} 
\label{tab:impute}
\centering
\begin{tabular}{cccc} 
 \toprule
 Case & Model & Average Imputation MSE & Average Offset MSE \\ 
 \midrule
 \multirow{5}{*}{MCAR} & BPCA   & 1.7856 &  0.0011\\
                       & PPCA   & 1.8170 &  0.0011\\
                       & Hurdle & 1.8195 &  0.0011\\
                       & NIPALS & 1.9105 &  0.0012\\ 
                       & Sample Mean & 4.6481 & 0.0012 \\ 
 \midrule
 \multirow{5}{*}{MAR}  & Hurdle & 1.8048 &  0.0020\\
                       & BPCA   & 1.8679 &  0.0034\\
                       & PPCA   & 1.8704 &  0.0028\\
                       & NIPALS & 2.2009 &  0.0081\\ 
                       & Sample Mean & 5.8782 & 0.0388\\ 
\bottomrule
\end{tabular}
\end{table}

The hurdle model representation provides several diagnostics for missing data which are not directly obtainable using other approaches. The first is based on the missingness probability score found using the sigmoid expression $1/\left[1 + \exp(- \bx_i\by_{1,1} - \mu_{1,1})\right]$ which represents the fitted value for the Boolean portion of the hurdle data. These scores can be used to construct ROC curves to measure how well the low-rank representation can discriminate between missing and non-missing occurrences. Figure \ref{fig:na} contains ROC curves for both the first simulated MCAR and MAR data sets. In the MCAR data sets, missingness is unexplained by the observed data and the resulting average area under the ROC curve (AUC) was 0.53. The MAR data sets had an average AUC of 0.88 which correctly suggests missingness is not likely to be completely at random. The interpretability of the AUC value is dependent on the degree of the low-rank model. Higher rank models which explain close to $100\%$ of the total loss lack interpretability since their representation will over-fit observed noise. In both MCAR and MAR cases the total loss reductions were near $80\%$ over the offset only models. This suggests missingness is difficult to explain for the MCAR example and remains as noise, whereas missingness is easily represented in the MAR case and does not remain as a contributor to unexplained loss. The latter finding suggests the hurdle model is a useful representation for the underlying MAR data structure. 

The second diagnostic is relevant when missingness is easily explained by the model. Variables associated with missingness can be identified by inspecting the cosine similarity between the vector $\by_{1,1}$ and the other columns $\by_j$ in $\bY$:   
\[
\theta_{j} = 1 - \frac{1}{\pi} \cos^{-1}\left[\frac{\by_{1,1} \cdot \by_j}{\|\by_{1,1}\| \|\by_j\|}\right].
\]
The cosine similarities can be converted into distances using $d_{j} = 1 - 2|\theta_{j} - 0.5|$, where $d_j \approx 0$ implies a high degree of dependence and $d_j \approx 1$ suggests no association. The similarities and distances for the first MAR data set are summarized in Table \ref{tab:dist}. The distance measures for columns $\by_2$ and $\by_3$ are small, which correctly suggests the values $a_{i2}$ and $a_{i3}$ are related to missingness. Interestingly columns $\by_9$ and $\by_8$ also report small distances. Upon inspection, simulated entries $a_{i9}$ and $a_{i8}$ were moderate to highly correlated with $a_{i2}$ and $a_{i3}$, indicating the reduced representation is distributing the observed influences across the collection of correlated variables. This outcome seems somewhat expected given the nature of low-rank models. Overall $87\%$ of the simulated MAR data sets had at least one of the two influential variables in the top two distance scores, and this increased to $100\%$ when considering the top three.

\begin{table}[h]
\caption{Variables associated with missingness}
\label{tab:dist}
\centering
\begin{tabular}{cccccccccccc}
\toprule
 & \multicolumn{11}{c}{Column} \\
  \cmidrule{2-12}
      & $\by_{1,1}$ & $\by_9$ & $\by_2$ & $\by_3$ & $\by_8$ & $\by_4$ & $\by_{1,2}$ & $\by_7$ & $\by_6$ & $\by_5$ & $\by_{10}$ \\
  \midrule
$d_j$        & 0 & 0.15 & 0.24 & 0.34 & 0.40 & 0.60 & 0.70 & 0.71 & 0.75 & 0.89 & 0.99 \\
$\theta_{j}$ & 1 & 0.08 & 0.12 & 0.17 & 0.80 & 0.30 & 0.35 & 0.65 & 0.37 & 0.55 & 0.50 \\
\bottomrule
\end{tabular}
\end{table}


\section{Summary}\label{sec:5}

This paper described the low-rank hurdle model which falls under the generalized low-rank framework. Previous authors have proposed the ZIFA model which is a special case of the reduced hurdle model. The methodology is particularly applicable to dimensionality reduction problems which exhibit characteristics similar to hurdle or zero-inflated regression problems. In addition to providing a more natural loss approximation, the hurdle model's design allows practitioners to examine aspects of the low-rank representation not readily available when using alternative procedures. This may be particularly useful in the case of missing data which was demonstrated in the applications.        

\end{document}